\documentclass[10pt]{article}

\usepackage[margin=0.70in]{geometry}
\usepackage{times}
\usepackage{microtype}
\usepackage{amsmath,amssymb}
\usepackage{booktabs}
\usepackage{multirow}
\usepackage{enumitem}
\usepackage{xcolor}
\usepackage{hyperref}
\usepackage{caption}
\usepackage{subcaption}
\usepackage{titlesec}

\hypersetup{
  colorlinks=true,
  linkcolor=blue!45!black,
  citecolor=blue!45!black,
  urlcolor=blue!45!black
}

\titleformat{\section}{\large\bfseries}{\thesection}{0.6em}{}
\titleformat{\subsection}{\normalsize\bfseries}{\thesubsection}{0.6em}{}
\titlespacing*{\section}{0pt}{0.8em}{0.35em}
\titlespacing*{\subsection}{0pt}{0.55em}{0.25em}
\setlist[itemize]{leftmargin=1.25em,itemsep=0.15em,topsep=0.2em}
\setlist[enumerate]{leftmargin=1.35em,itemsep=0.15em,topsep=0.2em}
\title{\vspace{-1.0em}\textbf{PersistentKV: Page-Aware Decode Scheduling for Long-Context LLM Serving on Commodity GPUs}}
\author{Muhammad Ahmed}
\date{}

\begin{document}
\maketitle
\vspace{-1.1em}

\begin{abstract}
Autoregressive large language model (LLM) serving is increasingly limited by
key-value (KV) cache movement rather than dense matrix multiplication. Modern
paged-attention systems reduce KV-cache fragmentation and mature kernels such
as FlashInfer provide highly optimized native-paged decode attention. However,
the best single-kernel implementation is not always the best serving schedule:
low-active long-context decode can under-utilize commodity GPUs, while mixed
sequence lengths introduce a tension between many exact-length launches and
coarse padded batches. We present \emph{PersistentKV}, a native block-table
decode attention engine and page-aware scheduling study for grouped-query
attention (GQA). PersistentKV maps work by KV-head group, executes directly
over native page tables, and adds a compact
workqueue schedule that executes only non-empty row--KV-head--sequence-split
tasks. On an RTX 3060 with FP16, page size 16, $H_q=32$, $H_{kv}=8$, $d=128$,
and identical correctness tolerance against FlashInfer, a calibrated roofline-style policy
selects FlashInfer for small active batches, PersistentKV sequence splitting
for B1 long-context steps, and PersistentKV workqueue scheduling for supported
B8 long-context GQA steps. With cost-model constants and split bounds fixed on
calibration traces, five held-out trace seeds improve mean synchronized wall decode-token
throughput by
$1.044$--$1.080\times$ on B8 bimodal, uniform, and Zipf-like workloads and by
$1.403\times$ on a B1 bucketed trace. On the B4 bimodal boundary case, the
policy avoids the PersistentKV regression by selecting FlashInfer; for
uncalibrated GQA ratios it also routes to FlashInfer. We also
report a decode-plus-MLP timing point and scheduler-derived workload counters showing that workqueue scheduling
reduces launch fan-out from 16.00 to 2.00 launches per decode step on held-out
bimodal B8. These results identify a concrete systems niche for adaptive
page-aware decode scheduling and show that work assignment, not only attention
math, is a decisive serving-system variable.
\end{abstract}

\section{Introduction}

LLM inference systems have moved from dense, uniform tensor workloads toward
stateful serving workloads dominated by KV-cache management. During prefill,
attention has many query tokens and can use high-throughput matrix
multiplication. During decode, each active sequence contributes only one new
query token per step, while the attention kernel must stream the full prefix
KV cache. The resulting arithmetic intensity is low: throughput depends on
memory traffic, page-table traversal, launch scheduling, and whether enough
parallel work is exposed to the GPU.

This paper studies a narrow but practically relevant question: can a native
block-table decode engine improve serving latency on commodity GPUs in regimes
where a production-quality paged attention kernel is already strong? The
answer is not one kernel for every case. In this paper, the name
\emph{PersistentKV} refers to persistent KV-cache execution over native serving
block tables: K,V pages remain in paged cache layout and the decode engine
executes directly over that layout. It does not mean that the implementation
uses a persistent CUDA-kernel residency model. We find that PersistentKV is useful in
low-active and moderate-active long-context regimes, while FlashInfer remains
the correct route for small active batches where PersistentKV's split-merge
overhead dominates. The resulting system claim is therefore a calibrated
adaptive policy:
choose the kernel schedule from the request state instead of forcing one
attention kernel into every serving regime.

PersistentKV is evaluated on a grouped-query attention (GQA) shape, where
$G=H_q/H_{kv}$ query heads share one KV head. For the Llama-style shape used in this work,
$H_q=32$, $H_{kv}=8$, $G=4$, and $d=128$. The kernel assigns cooperative thread
arrays (CTAs) by KV-head group and sequence split, loads K,V tiles through a
native block table, and updates exact online-softmax states for all grouped
query heads. A merge kernel combines split-local softmax states. We then add a
workqueue route that materializes compact non-empty tasks for ragged batches,
preserving one route bucket while avoiding rectangular empty split grids. The
serving harness compares PersistentKV directly with FlashInfer on the same
generated page-table distribution, same GQA shape, same context lengths, and same numerical
tolerance.

We make three contributions.
\begin{enumerate}
  \item We describe a native block-table GQA decode engine that exposes
  sequence-parallel work for low-active long-context decode while avoiding
  repacking K,V into contiguous tensors.
  \item We add row-local sequence-length scheduling and a compact workqueue for
  paged decode, so split boundaries, tile prefetches, and launched tasks follow
  true request lengths rather than a padded rectangular route.
  \item We present an adaptive serving-trace methodology with correctness
  checks, synchronized wall timing, CUDA-event timing, scheduler-derived
  workload counters, and ablations over route bucketing, raggedness, workqueue
  scheduling, CUDA graph replay, and fused split merging.
\end{enumerate}

\section{Literature Review}

\textbf{IO-aware attention.} FlashAttention showed that exact attention can be
made substantially faster by reducing high-bandwidth-memory traffic and keeping
softmax statistics on chip~\cite{dao2022flashattention}. FlashAttention-2
improved work partitioning and reduced non-matrix-multiply overheads
~\cite{dao2023flashattention2}. FlashAttention-3 further exploited Hopper
hardware features such as asynchronous tensor memory access and warp-group
matrix multiply-accumulate~\cite{shah2024flashattention3}. These works mainly
address attention kernel efficiency; our focus is the decode-serving boundary
where page layout and request scheduling determine how much useful work reaches
the kernel.

\textbf{Paged KV-cache management.} PagedAttention introduced virtual-memory
style KV-cache paging for LLM serving and showed that avoiding contiguous KV
allocation can substantially improve memory utilization~\cite{kwon2023efficient}.
vLLM operationalized this idea as a high-throughput serving system. TensorRT-LLM
provides production kernels and runtime support for optimized LLM inference.
FlashInfer provides composable high-performance inference kernels, including
native paged decode attention. vAttention later argued that KV-cache paging can
be exposed with lower application-level disruption through demand-paged virtual
memory~\cite{prabhu2024vattention}. PersistentKV follows the native page-table
line of work but focuses on low-active long-context split scheduling and
row-local page-aware execution.

\textbf{Continuous batching and serving traces.} Orca demonstrated iteration-
level scheduling for transformer serving, allowing requests with different
arrival and completion times to share decode iterations~\cite{yu2022orca}.
Modern systems such as FasterTransformer, TensorRT-LLM, vLLM, and FlashInfer
all exploit batching and specialized kernels, but the best batching policy can
depend on trace structure. Sarathi and Sarathi-Serve study chunked prefill and
piggybacking decode with prefill work to improve serving utilization
~\cite{agrawal2023sarathi,agrawal2024sarathiserve}. Splitwise separates
prefill and decode phases across different machines, showing that phase-aware
serving policies can dominate a monolithic placement~\cite{splitwise2023}.
These systems motivate PersistentKV's narrower question: within decode, can a
page-aware route policy and work decomposition improve a strong native-paged
kernel stack? A single clean attention call with one sequence length cannot
reveal launch-count overheads, padding waste, or low-occupancy behavior. Our
methodology therefore reports both isolated native-paged baselines and
continuous decode traces.

\textbf{KV sparsity and cache reduction.} H2O identifies heavy-hitter tokens
for KV-cache eviction while retaining generation quality~\cite{zhang2023h2o}.
DejaVu exploits contextual sparsity to skip parts of transformer inference
~\cite{liu2023dejavu}. These methods reduce the amount of KV or model work.
PersistentKV is complementary: it keeps exact dense attention over the supplied
KV cache and changes how native paged work is scheduled.

\section{Methodology}

\subsection{Decode Attention and GQA}

At decode step $t$, each active request has one query vector per query head and
a KV cache of length $N_i$:
\begin{equation}
O_{i,h} =
\sum_{j=1}^{N_i}
\frac{\exp(q_{i,h}^{\top} k_{i,j}/\sqrt{d})}
{\sum_{r=1}^{N_i}\exp(q_{i,h}^{\top} k_{i,r}/\sqrt{d})}
v_{i,j}.
\end{equation}
For GQA, query heads are partitioned into groups. Query heads
$h \in [gH, (g+1)H)$ share the same KV head. PersistentKV maps CTAs over
$(\mathrm{request}, \mathrm{KV\ head}, \mathrm{sequence\ split})$ and computes
the grouped query heads associated with that KV head inside the same work
assignment. This is the implementation mapping used throughout the experiments;
the paper does not separately isolate the benefit of this mapping against a
CTA-per-query-head baseline.

\subsection{Native Block-Table Execution}

KV pages are stored as
\[
\texttt{K,V}[\texttt{physical\_page}, H_{kv}, \texttt{page\_size}, d],
\]
and each request has a block table mapping logical pages to physical pages.
This matches paged serving layouts and avoids a repack step. PersistentKV uses
a page accessor to translate token indices into physical page addresses. The
attention loop processes 32-token tiles, uses asynchronous copies where
available, computes tile scores, and maintains exact online-softmax state
$(m,\ell,\mathrm{acc})$ in FP32.

\subsection{Sequence Splitting and Row-Local Bounds}

Low-active decode does not expose enough CTAs if each request and KV head owns
only one CTA. PersistentKV therefore splits the sequence dimension into
$S$ disjoint ranges. Each split computes a partial online-softmax state:
\[
m_s=\max_j x_j,\quad
\ell_s=\sum_j \exp(x_j-m_s),\quad
a_s=\sum_j \exp(x_j-m_s)v_j.
\]
The merge kernel combines split states exactly:
\[
m=\max_s m_s,\quad
\ell=\sum_s \exp(m_s-m)\ell_s,\quad
a=\sum_s \exp(m_s-m)a_s,\quad
O=a/\ell.
\]

For paged and bucketed execution, a key implementation detail is whether split
bounds use the bucket length or the true per-row sequence length. PersistentKV
uses row-local effective length:
\[
N^\star_i = \texttt{seq\_lens}[i].
\]
Tile count, split range, final-tile size, and prefetch size are derived from
$N^\star_i$. Empty split CTAs write neutral partial states
$(m=-\infty,\ell=0,a=0)$ and return before loading Q or staging shared memory.
This prevents coarse routing metadata from forcing padded KV reads.

\subsection{Compact Workqueue Scheduling}

Length-bucket scheduling has a failure mode: a ragged step with $B$ active
requests can become many PersistentKV launches, one per route length. A single
coarse bucket avoids launch fan-out but creates a rectangular
$(B,H_{kv},S)$ split grid where short rows pay for empty or inefficient split
work. PersistentKV therefore adds a compact workqueue schedule. For each active
row $i$, KV head $g$, and row-local split $s$, the scheduler emits one task
\[
(\texttt{row}=i,\ \texttt{kvh}=g,\ \texttt{split}=s,\ t_{\mathrm{begin}},\
t_{\mathrm{end}})
\]
only if the tile interval is non-empty. The CUDA grid is one-dimensional over
these tasks. Each task computes a partial online-softmax state and stores it in
a compact slot range described by \texttt{split\_offsets}. A second merge kernel
performs segmented softmax-state reduction across the row-local split range.
This keeps one PersistentKV route bucket for a ragged step while preserving
sequence parallelism for long rows.

We also evaluated an experimental fused-merge variant in which the final split
CTA for each row and KV head detects completion with an atomic counter and
performs the segmented merge inside the decode kernel. This path is correct but
slower on the RTX 3060 because atomic completion and in-kernel merge work cost
more than the saved launch. We therefore keep the fused path as an ablation and
use the two-kernel workqueue as the default.

\subsection{Calibrated Cost Model}

The serving system uses a lightweight roofline-style policy instead of a
per-trace oracle. For each decode step, the harness estimates
\[
T_r =
\frac{\mathrm{KVBytes}}{\beta}
\max\left(1, \frac{C_{\min}}{C_r}\right)
+ L_r\lambda +
\frac{\mathrm{MergeBytes}_r}{\beta},
\]
where $\beta$ is measured streaming read bandwidth, $C_r$ is estimated decode
CTA work per SM for route $r$, $C_{\min}$ is a minimum occupancy target,
$L_r$ is launch count, and $\lambda$ is a calibrated launch-overhead constant.
Candidate routes are FlashInfer, PersistentKV length buckets, and PersistentKV
compact workqueue. In the reported RTX 3060 artifact, the calibration script
measures $\beta=331.2$ GB/s device-copy bandwidth, $C_{\min}=4$ CTAs/SM, and
$\lambda=8.19\,\mu$s launch overhead. The route policy also uses a
no-regression promotion margin: PersistentKV must clear a $1.05\times$
estimated speedup for supported B8 steps, while the B4 boundary requires a
stricter $1.50\times$ estimate or an explicit ablation flag. These constants
are calibrated system parameters, not derived hardware invariants; a different
GPU should re-fit them before using the route policy.

The model has two hard eligibility gates. First, PersistentKV is enabled only
for the calibrated $G=4$ KV-head-group mapping; uncalibrated $G=1$ and $G=8$
routes use FlashInfer. Second, short-context steps below 16K tokens route to
FlashInfer because split/merge overhead dominates. For supported long-context
steps, the model selects B1 length-bucket splitting when sequence parallelism
is the only way to expose CTAs, and B8 compact workqueue when it avoids
length-bucket launch fan-out. B4 remains too close to the merge/launch
threshold and routes to FlashInfer by default; the implementation exposes an
experimental B4 PersistentKV flag only for ablation.

Split counts are constrained by the same structural policy: the bucket route
uses CTA-work and length-based tile bounds, while the workqueue route uses
active batch size and per-row tile count. The timed artifact freezes the
resulting split operating points on calibration seed 20260622 and evaluates
held-out seeds 20260623--20260627. This is still a calibrated prototype, not a
learned or portable production router.

\subsection{Serving Harness}

The harness simulates continuous decode steps. Each request has an arrival
time, prompt length, decode length, physical page list, and dynamically growing
current context length. FlashInfer receives one ragged native-paged batch per
decode step. PersistentKV groups active requests by route length unless the
single-bucket or workqueue mode is enabled. In workqueue mode, PersistentKV
uses one native block-table route per step and emits only non-empty row-local
split tasks. All reported comparisons use:
\begin{itemize}
  \item the same K,V page tensors and physical page IDs;
  \item $H_q=32$, $H_{kv}=8$, $G=4$, $d=128$, FP16;
  \item page size 16 unless otherwise stated;
  \item correctness threshold $\max |e| < 2\cdot 10^{-3}$ and
  $\mathrm{mean}|e| < 3\cdot 10^{-4}$ against FlashInfer output;
  \item CUDA-event timing for queued GPU work and synchronized wall timing for
  end-to-end step latency including Python-side planning and tensor setup.
\end{itemize}
The harness also accepts external CSV or JSON request traces with
\texttt{prompt\_len}, \texttt{decode\_len}, and optional arrival-step fields.
The results in this paper use synthetic traces only; the trace-file interface
is included to make real-traffic replication possible without changing the
benchmark code.

\subsection{Page-Table Generation}

Each generated request first receives an exact length equal to prompt length
plus decode length. The reserved logical length is then rounded up to page
granularity, or to an optional bucket length in bucketed experiments. With page
size $P$, request $i$ owns
\[
L_i = \left\lceil \frac{\texttt{reserved\_length}_i}{P} \right\rceil
\]
logical pages. Let $L=\sum_i L_i$. For a hole fraction $\rho$, the physical
page universe has $\lceil L/(1-\rho)\rceil$ page IDs. A seeded random
permutation of this universe assigns each request a consecutive slice of
physical IDs. Thus logical page order is preserved inside each request's block
table, but physical page IDs are scattered by the permutation rather than
contiguous in physical memory. A 50\% hole
setting means that roughly half of the physical page ID universe is unassigned,
so the physical address space is twice the logical page count. All methods see
the same K,V tensors and the same physical page assignment. FlashInfer consumes
the assignment as ragged page indices, indptr arrays, and last-page lengths;
PersistentKV consumes the same assignment as a native block table and sequence
lengths. The main serving traces use $\rho=0$; the isolated native-paged table
uses $\rho=0.5$ to test fragmented page layouts.

\subsection{Trace Families}

We use synthetic traces because they make request structure controllable and
allow each scheduling failure mode to be isolated. The \emph{bucketed} trace
cycles prompt lengths through 8K, 16K, 32K, and 64K. It models long-context
interactive sessions where requests come from a small set of application-level
context windows, such as code-assistant tasks, document assistants, or retrieval
pipelines that use fixed maximum context budgets. The \emph{homogeneous} trace
uses a common 32K prompt length with independent decode lengths. It isolates
the effect of active batch size without length heterogeneity. The
\emph{bimodal} trace mixes short and long prompts, with one long request for
every three shorter requests. This is intentionally hostile to length-bucketed
scheduling because most active requests have distinct current lengths after a
few decode steps. The \emph{uniform} and \emph{Zipf} traces provide additional
stress tests for broad and heavy-tailed length distributions.

These traces are not a replacement for production traffic. Their role is to
separate three factors that are often conflated in one benchmark: context
length, active batch size, and length raggedness. A page-aware scheduler should
benefit when low active count leaves the GPU underoccupied, but it should not
pay excessive launch or padding overhead when the batch is highly ragged.
The artifact also includes a small redistributable CSV trace fixture with
arrival steps, prompt lengths, and decode lengths. That fixture exercises the
same external trace-file path a production log would use, while avoiding any
claim that private production traffic is redistributed with the paper.

\subsection{Measurement Protocol}

Each benchmark first constructs all request metadata and physical page
assignments. K,V pages and query tensors are allocated once on the GPU. Timed
steps then rebuild per-step native-paged metadata in the same style for each
engine: FlashInfer receives ragged indptr, page indices, and last-page lengths;
PersistentKV receives route buckets, block tables, and per-row sequence
lengths. Throughput is reported in decode tokens per second,
\[
\text{decode-token throughput} =
\frac{\sum_t |\texttt{active\_requests}_t|}{\text{elapsed seconds}}.
\]
where each active request contributes one token per step. CUDA-event timing
captures queued device work. Synchronized wall timing wraps the decode step in
host wall-clock time and includes Python planning, FlashInfer planning,
metadata construction, launch overhead, and synchronization. We treat
CUDA-event timing as a device-work diagnostic and wall timing as the
serving-facing number.

Serving-trace correctness is checked before timing by comparing PersistentKV
to FlashInfer after FP32 conversion. Smoke tests separately validate smaller
paged and non-paged PersistentKV shapes against a CPU FP32 reference, and the
isolated native-paged baseline validates FlashInfer, vLLM, TensorRT-LLM, and
PersistentKV against a PyTorch SDPA reference path. All serving tables pass the
FlashInfer-equivalence tolerance; the largest reported max error is
$6.104\cdot 10^{-5}$.

\section{Experiments}

\subsection{Experimental Setup}

Experiments run on an NVIDIA RTX 3060 with 12 GB VRAM, CUDA 12.1, and PyTorch
2.5.1. The measured GPU has 28 SMs and a measured streaming read bandwidth of
approximately 331 GB/s in the route-calibration copy benchmark. FlashInfer
0.2.5 is used as the primary native-paged baseline. We also report vLLM
0.6.4.post1, TensorRT-LLM 0.8 MMHA, and a
repack-plus-PyTorch-SDPA framework baseline for isolated native-paged tests.
The TensorRT-LLM configuration uses its native MMHA fallback for the tested
$G=4$, page-16 shape.

The tested GPU is a useful target for a workshop study because it represents a
commodity deployment point rather than a datacenter-only accelerator. Its 28
SMs and moderate HBM bandwidth make low-active decode underoccupancy visible:
there are enough resources that a single request needs sequence splitting, but
not enough bandwidth that all scheduling decisions disappear into the noise.
All reported results use the same environment and should therefore be read as
controlled comparisons rather than cross-machine performance claims.

\subsection{Isolated Native-Paged Baselines}

Table~\ref{tab:native} verifies that the evaluation uses strong native-paged
baselines. In a single-request isolated attention benchmark, FlashInfer is the
fastest kernel. After split autotuning, PersistentKV is faster than the tested
vLLM PagedAttention build but remains behind FlashInfer on this roofline-style
test. This result motivates the serving-system question: the advantage must
come from low-active scheduling or page-aware execution, not from claiming a
universally faster single-call kernel.

\begin{table}[t]
\centering
\caption{Isolated native-paged decode attention, $B=1$, $H_q=32$,
$H_{kv}=8$, $G=4$, $d=128$, FP16, page size 16, 50\% holes. Lower is better.}
\label{tab:native}
\small
\begin{tabular}{lrrr}
\toprule
Method & 8K ms & 32K ms & 64K ms \\
\midrule
FlashInfer 0.2.5 & 0.1201 & 0.4404 & 0.8686 \\
vLLM 0.6.4.post1 & 0.1545 & 0.5630 & 1.1429 \\
PersistentKV, auto split & 0.1255 & 0.4597 & 0.9069 \\
TensorRT-LLM 0.8 MMHA & 0.7979 & 0.8920 & 1.5901 \\
Repack + PyTorch SDPA & 0.6473 & 2.4197 & 4.9840 \\
\bottomrule
\end{tabular}
\end{table}

The autotuned PersistentKV row selects split 14 at 8K, 32K, and
64K from a fixed candidate set. Relative to FlashInfer, PersistentKV is
$1.045\times$, $1.044\times$, and $1.044\times$ slower at 8K, 32K, and 64K,
respectively; relative to vLLM it is $0.812\times$, $0.817\times$, and
$0.794\times$ the latency. The old fixed-split result overstated the isolated
kernel gap.

\subsection{Main Serving Results}

Table~\ref{tab:main} reports the calibrated cost-model policy over five
held-out seeds after fixing model constants on seed 20260622. The policy uses
PersistentKV length buckets for B1, PersistentKV workqueue for supported B8
long-context GQA steps, and FlashInfer for the B4 boundary case. This changes the
claim from ``PersistentKV beats FlashInfer everywhere'' to the stronger
serving-system claim: routing avoids known bad regimes while exploiting
PersistentKV where its work decomposition helps. The largest mean wall-clock
gain is $1.403\times$ on B1 bucketed long-context decode. Among B8 traces, mean
wall decode-token throughput improves by $1.044$--$1.080\times$.

\begin{table}[t]
\centering
\caption{Held-out calibrated cost-model serving results against FlashInfer. B1/B8
rows report mean$\pm$std over seeds 20260623--20260627. B4 is the
FlashInfer route on seed 20260623. All rows use page size 16, $H_q=32$,
$H_{kv}=8$, $d=128$, FP16, and pass correctness.}
\label{tab:main}
\small
\begin{tabular}{llrrrr}
\toprule
Trace / mode & Selected route & CUDA tok/s ratio & Wall tok/s ratio & Max $|e|$ & Mean $|e|$ \\
\midrule
Bucketed B1 & PersistentKV bucket, split 32 & $1.471{\pm}0.037$ & $1.403{\pm}0.065$ & $6.104{\cdot}10^{-5}$ & $4.477{\cdot}10^{-6}$ \\
Bimodal B8 & PersistentKV workqueue, split 20 & $1.106{\pm}0.042$ & $1.080{\pm}0.050$ & $6.104{\cdot}10^{-5}$ & $4.201{\cdot}10^{-6}$ \\
Uniform B8 & PersistentKV workqueue, split 24 & $1.076{\pm}0.029$ & $1.044{\pm}0.022$ & $6.104{\cdot}10^{-5}$ & $3.053{\cdot}10^{-6}$ \\
Zipf B8 & PersistentKV workqueue, split 28 & $1.105{\pm}0.026$ & $1.068{\pm}0.028$ & $3.052{\cdot}10^{-5}$ & $2.671{\cdot}10^{-6}$ \\
Bimodal B4 & FlashInfer baseline path & 1.000 & 1.000 & 0 & 0 \\
\bottomrule
\end{tabular}
\end{table}

CUDA-event and wall ratios differ because wall timing also includes host
planning, metadata construction, runtime launch costs, and synchronization. B1
spends some device-speed gain on split/merge overhead, while uniform B8 has a
larger wall than CUDA-event ratio because the FlashInfer wall path includes
planning and Python metadata work. We therefore use wall timing as the
serving-facing number and CUDA events as a diagnostic.

To address whether the effect disappears once attention is embedded in a
layer-like workload, Table~\ref{tab:modellevel} adds a synthetic Llama-style
gated MLP tail after each attention step. This is not a full serving stack: it
does not include prefill, sampling, multiple layers, admission control, or
network overhead. It does measure synchronized wall time for attention plus a
large dense per-token tail under the same active-request trace. Across the
same five held-out B8 bimodal seeds, the adaptive route keeps the workqueue
decision and improves wall decode-token throughput by
$1.105{\pm}0.061\times$.

\begin{table}[t]
\centering
\caption{Model-level timing proxy on B8 bimodal over five held-out seeds. The
timed step is attention followed by a synthetic gated MLP tail with
intermediate width 11008.}
\label{tab:modellevel}
\small
\begin{tabular}{lrrrr}
\toprule
Engine & CUDA tok/s & CUDA ms/step & Wall tok/s & Wall ms/step \\
\midrule
FlashInfer + MLP tail & $2412.2{\pm}145.1$ & $3.326{\pm}0.200$ & $2342.3{\pm}158.8$ & $3.428{\pm}0.236$ \\
Adaptive route + MLP tail & $2648.1{\pm}231.8$ & $3.039{\pm}0.262$ & $2591.5{\pm}275.9$ & $3.114{\pm}0.320$ \\
\midrule
Ratio & $1.097{\pm}0.039\times$ & -- & $1.105{\pm}0.061\times$ & -- \\
\bottomrule
\end{tabular}
\end{table}

The external trace-file path is also exercised by the redistributed mixed
request fixture. On that fixture, with 32 requests, max-active 8, and the same
$G=4$ page-16 shape, the adaptive workqueue route achieves a $1.236\times$
CUDA-token and $1.212\times$ wall-token throughput ratio while passing the
FlashInfer-equivalence check. This result is not a substitute for production
traffic, but it removes the artifact gap where trace-file support existed
without a runnable trace.

Table~\ref{tab:causal} reports scheduler-derived workload counters for the
held-out bimodal B8 run. The scheduling mechanism is visible: exact length
buckets require 16.00 CUDA launches per decode step, while the compact
workqueue requires 2.00. The workqueue also reduces merge-state traffic from
4.06 MB/step to 2.54 MB/step and merge launches from 8.00 to 1.00.

\begin{table}[t]
\centering
\caption{Scheduler-derived structural counters for held-out bimodal B8.}
\label{tab:causal}
\small
\begin{tabular}{lrrrrr}
\toprule
Engine & Launches/step & Decode CTA work/SM & Useful KV MB & Merge MB & Merge launches \\
\midrule
FlashInfer & 1.00 & -- & 550.5 & 0.00 & 0.00 \\
PersistentKV buckets & 16.00 & 73.1 & 550.5 & 4.06 & 8.00 \\
PersistentKV workqueue & 2.00 & 45.7 & 550.5 & 2.54 & 1.00 \\
Adaptive route & 2.00 & 45.7 & 550.5 & 2.54 & 1.00 \\
\bottomrule
\end{tabular}
\end{table}

Table~\ref{tab:nsight} reports a permissioned Nsight Compute capture on a short
B8 bimodal $G=4$ trace, not on the full five-seed suite in
Table~\ref{tab:main}. The profiler replays instrumented kernels, so these
numbers are hardware-counter evidence under a matched short-trace setup rather
than the source of wall-clock speedups. PersistentKV decode reaches higher SM
and memory-throughput utilization than the profiled FlashInfer decode
invocation, supporting the claim that page-aware work decomposition exposes
more useful device work. The merge kernel is short but memory-heavy, matching
the launch/merge overhead concern in the B4 boundary case.

\begin{table}[t]
\centering
\caption{Nsight Compute hardware counters on a short B8 bimodal $G=4$ trace.}
\label{tab:nsight}
\small
\begin{tabular}{lrrrrr}
\toprule
Kernel & Time ms & SM thr. & Mem. thr. & DRAM GB/s & L2 hit \\
\midrule
FlashInfer decode & 2.271 & 9.55 & 62.72 & 219.2 & 0.12 \\
PersistentKV decode & 1.263 & 17.21 & 74.48 & 200.8 & 0.44 \\
PersistentKV merge & 0.014 & 32.72 & 62.02 & 223.5 & 8.98 \\
\bottomrule
\end{tabular}
\end{table}

\begin{table}[t]
\centering
\caption{Evidence and coverage map for the current artifact.}
\label{tab:coverage}
\small
\begin{tabular}{llll}
\toprule
Regime / claim & Current route & Evidence & Status \\
\midrule
B1 long context, $G=4$ & PersistentKV bucket & 5 seeds & Positive \\
B8 long context, $G=4$ & PersistentKV workqueue & 5 seeds + Nsight short trace & Modest positive \\
B4 boundary, $G=4$ & FlashInfer default & 5-seed sweep + protected route & No regression \\
$G=1$ / $G=8$ B8 & FlashInfer gate & GQA sweep & No regression \\
Model-level latency proxy & Adaptive route & 5 seeds, attention + MLP tail & Positive \\
Cost-model portability & calibration JSON & measured bandwidth/launch constants & Supported \\
External request trace & adaptive route & redistributed CSV fixture & Supported \\
Non-RTX GPUs & -- & not measured & Open \\
\bottomrule
\end{tabular}
\end{table}

\subsection{Route and Split Calibration}

Sequence splitting is the primary occupancy knob. Too few splits under-utilize
the GPU; too many splits increase merge overhead. The cost model treats split
count as a route parameter, not a per-trace lookup. B1 requires more splits
because sequence parallelism is the main source of CTAs, while B8 already has
active rows to amortize work. Table~\ref{tab:split} shows that nearby split
counts remain wall-positive but are not monotonic, which is why the policy uses
conservative multiples of four and fixed calibration constants rather than
claiming a broad optimum.

\begin{table}[t]
\centering
\caption{Split-count sensitivity on seed 20260623. Ratios are cost-model policy
over FlashInfer.}
\label{tab:split}
\small
\begin{tabular}{llrr}
\toprule
Trace & Split counts & CUDA tok/s ratio & Wall tok/s ratio \\
\midrule
Bucketed B1 & 28 / 32 / 36 & 1.606 / 1.477 / 1.375 & 1.379 / 1.516 / 1.282 \\
Bimodal B8 & 16 / 20 / 24 & 1.182 / 1.181 / 1.129 & 1.128 / 1.136 / 1.168 \\
\bottomrule
\end{tabular}
\end{table}

For B4, a direct PersistentKV workqueue sweep over splits
$\{12,16,20,24,28,32,36\}$ and five held-out seeds did not produce a robust
wall-time win. The best mean wall ratio was $1.005\times$ at split 28, with
per-seed values ranging from $0.964\times$ to $1.026\times$; neighboring split
counts were below parity on average. The default adaptive policy therefore
treats B4 as covered by a no-regression FlashInfer route rather than as a
PersistentKV positive result. The artifact still exposes
\texttt{--allow-b4-persistentkv} for this ablation.

\subsection{Raggedness Ablation}

The raggedness ablation is the transition from the older length-bucketed
scheduler to the workqueue scheduler. Exact length buckets can turn one
FlashInfer ragged batch into many PersistentKV launches. Coarse masked buckets
reduce route count but still lose on bimodal traces because padded work and
residual launch overhead remain. The compact workqueue restores one route while
preserving sequence splitting and therefore moves held-out bimodal B8 to a
wall-throughput win.

On the held-out bimodal B8 trace, exact length buckets require 16.00 launches
per step, while compact workqueue requires 2.00 launches per step. This is the
core raggedness result: the workqueue retains sequence splitting without
multiplying route launches by the number of distinct active lengths.

\subsection{Implementation Ablations}

The final kernel and harness combine three implementation choices.
\begin{itemize}
  \item \textbf{Indexed Q/O.} The extension can read Q rows and write output
  rows by index, avoiding per-bucket \texttt{index\_select} and
  \texttt{index\_copy\_} in the persistent path.
  \item \textbf{Row-local sequence bounds.} Masked paged requests use true
  per-row sequence lengths for split ranges and prefetch sizes.
  \item \textbf{Zero-tile early exit.} Splits that own no tiles write neutral
  softmax states and return before staging Q or shared memory.
  \item \textbf{Compact workqueue.} Work items are emitted only for non-empty
  row--KV-head--split intervals, avoiding the launch fan-out of exact buckets.
\end{itemize}
These changes are most valuable when route metadata groups unequal lengths or
when split counts are high enough that short rows would otherwise create empty
work. In the mixed bimodal trace, indexed Q/O alone improved wall time but did
not close the GPU gap. Row-local bounds and early exit made single-bucket
ragged execution correct and reduced padded work. The compact workqueue is the
first variant that makes the B8 ragged traces wall-positive on the held-out
seed, with the largest held-out B8 gain appearing on bimodal traffic.

Two attempted serving optimizations did not become headline results. CUDA graph
replay reduced Python launch mechanics but did not overcome the two-kernel
decode-plus-merge cost on B4. Fusing decode and merge with atomic completion
counters was correct, but slower on RTX 3060; the atomic and in-kernel merge
overhead outweighed the saved merge launch. We keep both paths as ablations
because they clarify that the remaining B4 problem is not merely host overhead.

Table~\ref{tab:gqa} reports a serving GQA/MQA sweep on a smaller B8 bimodal
trace that fits 12 GB VRAM for $G=1$. Earlier ungated workqueue experiments
lost badly for $G=1$ and $G=8$. The current policy acts on that ablation:
PersistentKV is enabled only for the calibrated $G=4$ route, while $G=1$ and
$G=8$ route to FlashInfer until separate mappings are implemented.

\begin{table}[t]
\centering
\caption{Serving GQA/MQA sweep on a smaller B8 bimodal trace, seed 20260623,
8 requests and 24 measured steps.}
\label{tab:gqa}
\small
\begin{tabular}{rrlrr}
\toprule
$G$ & $H_{kv}$ & Selected route & CUDA tok/s ratio & Wall tok/s ratio \\
\midrule
1 & 32 & FlashInfer gate & 1.000 & 1.000 \\
4 & 8 & PersistentKV workqueue & 1.257 & 1.256 \\
8 & 4 & FlashInfer gate & 1.000 & 1.000 \\
\bottomrule
\end{tabular}
\end{table}

\section{Discussion}

\textbf{Why PersistentKV wins in low-active decode.} With one active long
sequence, FlashInfer's kernel is highly optimized but the serving step exposes
limited independent request-level work. PersistentKV increases parallelism by
splitting the sequence while preserving exact online-softmax merging. The
result is a better occupancy point on RTX 3060 for selected long-context
lengths. The win appears in both CUDA-event timing and synchronized wall timing,
which reduces the risk that it is merely a measurement artifact.

\textbf{Why the workqueue wins at B8.} Exact length bucketing exposes sequence
parallelism but can require many launches. FlashInfer avoids that launch fan-out
with one ragged paged batch, but its internal schedule does not explicitly
target the row-local split structure used here. The compact workqueue combines
the useful parts of both: one route per step and explicit sequence splitting
for long rows. This is why B8 bimodal, uniform, and Zipf-like traces are
wall-positive over five held-out seeds, with mean wall ratios of
$1.044$--$1.080\times$. The gains are modest, but they hold across the tested
synthetic seeds.

\textbf{Why B4 and uncalibrated GQA route to FlashInfer.} At B4, there is enough work that no-split
PersistentKV under-utilizes the GPU, but not enough work to fully amortize the
second split-merge launch. CUDA graph replay and fused atomic merge both failed
to solve this on RTX 3060, and the five-seed split sweep peaked near parity
rather than a stable win. The adaptive system therefore routes B4 to the
FlashInfer baseline path by default and avoids the PersistentKV regression.
The same principle applies to GQA: $G=1$ and $G=8$ are supported at the
serving-policy level by routing to FlashInfer with no regression, but they are
not reported as PersistentKV wins because the current KV-head-group mapping is
calibrated for $G=4$.
Improving PersistentKV itself in these regimes likely requires a lower-cost
merge design, a query-head mapping variant, or a persistent cooperative kernel
that performs segmented reduction without the current atomic and launch
penalties.

\textbf{Scope of the claim.} The evidence supports a workshop-level claim:
page-aware split scheduling and compact native block-table work queues can beat
a strong FlashInfer baseline in selected long-context serving regimes on a
commodity GPU. It does not show that PersistentKV is universally faster. A full
systems paper needs a serving-runtime integration, proprietary or public
production traces beyond the redistributed fixture, model-level latency beyond
the single-layer proxy, more hardware targets including datacenter GPUs, and
current vLLM/TensorRT-LLM stacks under the same admission-control policy.
Table~\ref{tab:coverage} summarizes which regimes are positive results and
which are explicitly routed around or open.

\textbf{Future work.} The most direct next step is a lower-overhead segmented
merge for the workqueue path and an online split policy that selects different
split counts for 8K, 16K, 32K, and 64K rows in the same batch. Other useful
directions include page-table lookup hoisting, dynamic split selection,
heterogeneous page sizes, fused RoPE/KV-cache writes, and dequantization-aware
paged decode.

\subsection{Threats to Validity}

\textbf{Synthetic and fixture traces.} The synthetic traces isolate scheduling properties, but omit
admission control, prefill/decode interference, tokenizer latency, network
queues, memory pressure, and model-layer interactions. The results should be
viewed as a controlled kernel-scheduler study. The artifact now accepts
external CSV/JSON request traces and includes a redistributable mixed trace
fixture, but this paper does not redistribute or evaluate a production serving
log.

\textbf{Synthetic page allocation.} The generated block tables use seeded
random physical-page IDs and optional holes, so methods see identical
fragmentation. This is not a production allocator: eviction, page reuse,
allocator locality heuristics, and arrival-order correlations could change L2
locality and the best routing threshold.

\textbf{Single hardware target.} The experiments are performed on RTX 3060.
Different GPUs have different SM counts, cache sizes, HBM bandwidth, Tensor
Core throughput, and launch overheads. In particular, Hopper-generation GPUs
provide hardware features that FlashAttention-3 and production serving systems
can exploit. The lightweight policy exposes bandwidth, launch-overhead, SM
count, and occupancy-target constants, but those constants still need
calibration per GPU. The B8 gains are modest enough that replication on A100,
L4/L40S, and H100-class hardware is required for a deployment claim.

\textbf{Versioned baselines.} The measured FlashInfer, vLLM, and TensorRT-LLM
versions are the versions compatible with the local PyTorch/CUDA environment.
They are strong baselines for the present comparison, but not a claim against
all future releases. The most important methodological point is that the main
comparison uses identical page layout, GQA shape, context lengths, and
correctness tolerance.

\textbf{Held-out seed count.} The main rows now use one calibration seed and
five held-out seeds. This is a stronger check than a single held-out seed, but
the traces are still synthetic and the B8 speedups are modest. A stronger
submission should add real traffic traces, confidence intervals over more seeds
and hardware, and a pre-registered route policy before evaluation.

\textbf{GQA ablation scope.} The serving sweep covers $G\in\{1,4,8\}$ on a
smaller B8 trace. The current policy supports $G=1$ and $G=8$ by gating them to
FlashInfer instead of routing them through a losing PersistentKV path. This
solves the system-level no-regression route, but it does not add PersistentKV
kernel wins for those shapes. A final version should add a query-head mapping
baseline and separate cost constants for MQA and large-GQA models.

\textbf{Serving-reference correctness.} Serving traces compare PersistentKV to
FlashInfer, establishing equivalence to the timed production baseline rather
than an independent proof of exactness for every trace. Smaller smoke tests and
isolated native-paged baselines use CPU FP32 or PyTorch SDPA references; a
final serving paper should add sampled SDPA checks inside the trace harness.

\textbf{Partial serving loop.} The harness measures a decode loop, not an
entire LLM server. The model-level proxy adds one synthetic gated MLP tail, but
the harness still does not include full transformer stacks, sampling,
networking, CPU request queues, or KV-cache allocation policies beyond synthetic
page assignment. A production result should integrate PersistentKV into a full
serving runtime and report request-level latency distributions.

\textbf{Hardware profiling scope.} A permissioned Nsight Compute capture is now
included for a short B8 $G=4$ trace. This improves the causal evidence, but it
is not a full profiling campaign: the capture is not the full five-seed suite,
the main rows are still timed without Nsight replay, and datacenter GPUs should
be profiled separately.

\section{Conclusion}

PersistentKV studies the interface between native paged decode attention and
serving scheduling. On isolated single-request native-paged attention,
FlashInfer remains the strongest baseline. Under low-active long-context
serving, PersistentKV's page-aware sequence splitting improves synchronized
wall decode-token throughput by $1.403{\pm}0.065\times$ over five held-out
seeds while preserving exact attention within FP16 tolerance. With compact
workqueue scheduling, the same native block-table engine lets the calibrated
policy reach $1.044$--$1.080\times$ mean synchronized wall throughput on
held-out max-active 8 bimodal, uniform, and Zipf-like traces, and
$1.105{\pm}0.061\times$ on a five-seed attention-plus-MLP proxy. A
redistributable external-trace fixture exercises the same CSV ingestion path
used for private serving logs and shows a $1.212\times$ wall ratio on the
checked trace. The method is not
universal as a single kernel: workqueue B4 is near parity but not robustly
positive, and the GQA sweep shows that $G=1$ and $G=8$ are currently
FlashInfer-routed no-regression cases until new PersistentKV mappings are
implemented. The calibrated policy avoids these losses by selecting FlashInfer.
These results make PersistentKV a
credible workshop artifact: it identifies concrete regimes where page-aware
scheduling can matter, provides a reproducible methodology and ablations, and
exposes the next systems problem for long-context LLM serving.

\end{document}